\documentclass[runningheads]{llncs}
\usepackage{booktabs}
\usepackage{multirow}
\usepackage[T1]{fontenc}
\usepackage{graphicx,verbatim}
\usepackage{hyperref} 
\usepackage{color}

\usepackage[table]{xcolor}
\usepackage{amsmath, amssymb}

\begin{document}
%
\title{T2I-Diff: fMRI Signal Generation via Time-Frequency Image Transform and Classifier-Free Denoising Diffusion Models}
\titlerunning{fMRI Generation via Time-Frequency Transform and Diffusion Models}
\authorrunning{H. H. Tew, et al.}
%

\author{Hwa Hui Tew\index{Tew, Hwa Hui}\inst{1} \and
Junn Yong Loo\index{Loo, Junn Yong}\inst{1}\textsuperscript{*} \and
Yee-Fan Tan\index{Tan, Yee-Fan} \inst{1} \and
Xinyu Tang\index{Tang, Xinyu} \inst{1} \and
Hernando Ombao\index{Ombao, Hernando} \inst{2} \and
Fuad Noman\index{Noman, Fuad} \inst{1} \and
Raphaël C.-W. Phan\index{Phan, Raphaël C.-W.} \inst{1} \and
Chee-Ming Ting\index{Ting, Chee-Ming}\inst{1}\textsuperscript{*}}

\institute{School of Information Technology, Monash University Malaysia \and
Statistics Program, King Abdullah University of Science and Technology
\email{\{loo.junnyong,ting.cheeming\}@monash.edu}}


\maketitle              
\renewcommand{\thefootnote}{\fnsymbol{footnote}}
\footnotetext[1]{Corresponding authors}
\begin{abstract}
Functional Magnetic Resonance Imaging (fMRI) is an advanced neuroimaging method that enables in-depth analysis of brain activity by measuring dynamic changes in the blood oxygenation level-dependent (BOLD) signals. However, the resource-intensive nature of fMRI data acquisition limits the availability of high-fidelity samples required for data-driven brain analysis models. While modern generative models can synthesize fMRI data, they often underperform because they overlook the complex non-stationarity and nonlinear BOLD dynamics. To address these challenges, we introduce T2I-Diff, an fMRI generation framework that leverages time-frequency representation of BOLD signals and classifier-free denoising diffusion. Specifically, our framework first converts BOLD signals into windowed spectrograms via a time-dependent Fourier transform, capturing both the underlying temporal dynamics and spectral evolution. Subsequently, a classifier-free diffusion model is trained to generate class-conditioned frequency spectrograms, which are then reverted to BOLD signals via inverse Fourier transforms. Finally, we validate the efficacy of our approach by demonstrating improved accuracy and generalization in downstream fMRI-based brain network classification. The code is available at \href{https://github.com/htew0001/T2I-Diff.git}{repository}

\keywords{fMRI  \and Time-Frequency Image \and Diffusion Models.}

\end{abstract}
%
%
%
\section{Introduction}
With the rapid advancement of artificial intelligence, deep generative modeling has shown promising capability in generating realistic variations of neuroimaging data \cite{IJCAI2024}. Although the generation of brain connectivity has been extensively studied, research on the direct generation of fMRI signals remains limited \cite{survey1,survey2}. fMRI signals are particularly valuable as they provide critical insights into neural activity, enabling more precise assessment and identification of neuropsychiatric and neurodevelopmental disorders \cite{ICIP2022,JBHI2024}. Despite their significance, acquiring fMRI signals is challenging, often resulting in small sample sizes that limit the performance of data-driven brain analysis models \cite{limitations3}. These limitations can lead to data imbalances and a lack of temporal dynamics in acquired brain signals, ultimately affecting the accuracy of predictive and diagnostic models for neurological and psychiatric conditions \cite{limitations1,TMI2022}. To address these challenges, generative techniques have been explored for synthesizing fMRI signals, enabling data augmentation to enhance various downstream applications \cite{limitations2,ICIP2023}.

Generative models, such as Generative Adversarial Networks (GANs) and Variational Autoencoders (VAEs), have been widely used to synthesize realistic time series data \cite{surveyGM}. While GANs are known for generating high-quality samples, VAEs are favored for their faster generation speed. However, both models face challenges in achieving stable training, often struggling with issues like mode collapse and optimization difficulties. For instance, Yoon et al. proposes TimeGAN, which jointly trains adversarial and supervised loss components within a learned embedding space, successfully preserving both the static and dynamic characteristics of synthetic time-series data \cite{timegan}. Apart from that, TimeVAE incorporates time-series components into its encoder-decoder network, enhancing interpretability in time-series generation, where it has demonstrated success in reducing overall training time compared to adversarial methods \cite{timevae}. 

Recently, denoising diffusion probabilistic models (DDPMs) have gained traction in time-series applications due to their robustness against model collapse \cite{surveyTS}. However, their potential for capturing the complex intrinsic properties of brain signals remains underexplored \cite{survey1,survey2}. For example, Coletta et al. propose different types of constraints to improve time-series generation: hard constraints enforce fixed points and global minima; soft constraints introduce penalties to guide the model towards desired temporal trends \cite{difftime}. Furthermore, ImagenTime leveraged the time-frequency images to generate synthetic data for general time-series benchmarks, demonstrating the feasibility of capturing intricate spectral and temporal patterns \cite{naiman2024utilizing}. The recently proposed Diffusion-TS model introduces a novel non-autoregressive diffusion model for time-series data, including fMRI signals. This framework explicitly captures temporal dynamics through transformer-based model architecture and disentangled seasonal-trend representations, achieving success across various generation tasks \cite{diffusionts}.

To address the challenge of insufficient inductive bias in dynamic brain characterization, this paper introduces T2I-Diff, an fMRI generation framework that leverages time-frequency image representations of BOLD signals and classifier-free denoising diffusion. This approach effectively captures both spectral and dynamic features underlying fMRI BOLD signals, which are essential for downstream brain disorder classification.
Additionally, our framework incorporates classifier-free denoising diffusion and EDM sampling for high-fideliy generation of time-windowed fMRI spectrograms conditioned on subject classes. The generated time-frequency images are then reverted to BOLD signals via image-to-time series transforms. Our method eliminates the need for additional classifiers to guide brain profiles, thereby reducing training redundancy and simplifying model optimization. We summarized our main contributions as follows:
\begin{enumerate}
    \item Our proposed T2I-Diff framework is the first to integrate a time-frequency image transform to capture complex spatiotemporal and spectral features for fMRI BOLD signal generation and brain disorder classification.
    \item We introduce classifier-free denoising diffusion and EDM sampling to generate conditioned time-windowed fMRI spectrograms without requiring an additional classifier. This enables T2I-Diff to dynamically modulates the diffusion process to guide the model in learning different brain signal profiles.
    \item Our results shows that T2I-Diff demonstrates competitive performance on time-frequency image generation and brain disorder (MDD) classification complemented by class-conditioned synthetic fMRI BOLD signals.
\end{enumerate}

\begin{figure}[t]
\includegraphics[width=\linewidth]{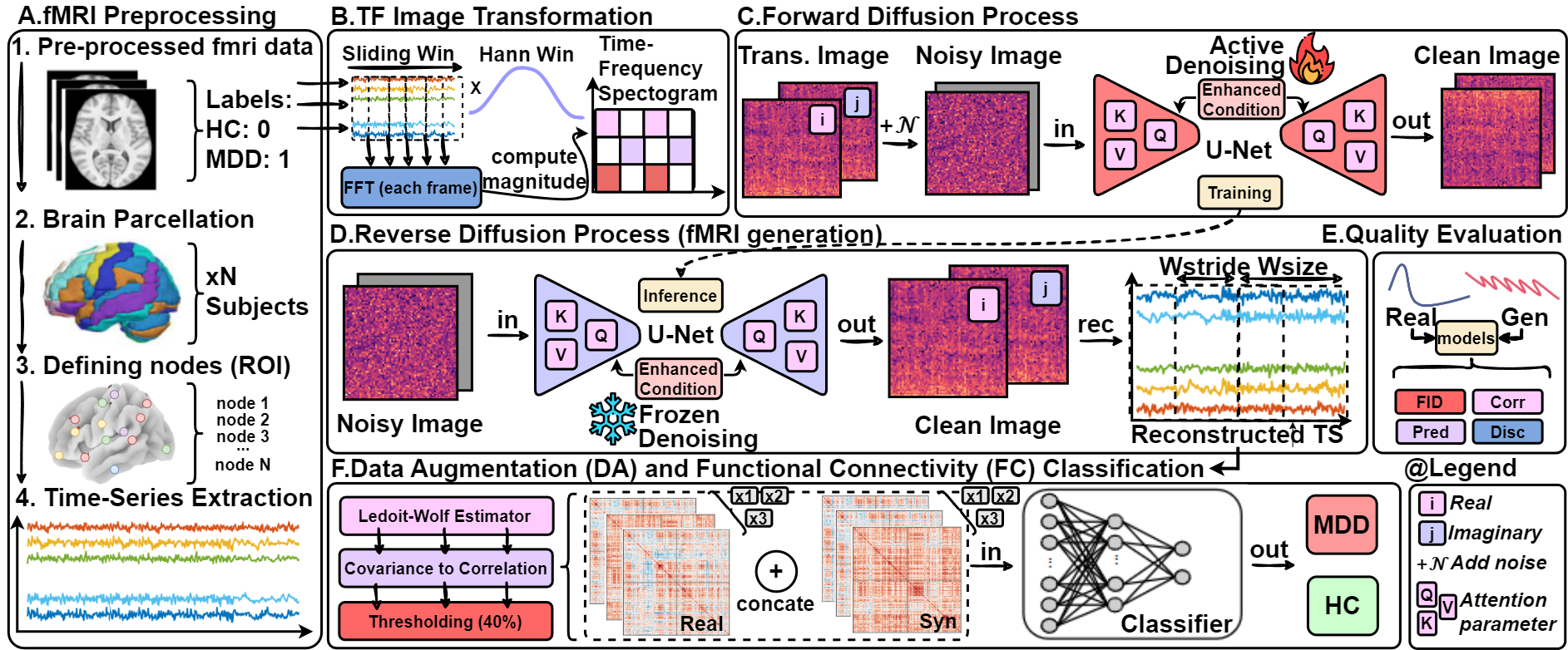}
\caption{Our framework T2I-Diff begins by (A.) preprocessing raw fMRI data in time domain and transforming into (B.) time-frequency spectrogram (image) using windowed Fourier transform. Next, the images are fed into the diffusion model executes both (C.) forward and (D.) reverse processes. Finally, we perform (E.) quality evaluation of the generation and apply (F.) DA on top of the real fMRI signals for MDD classification.} 
\label{fig1}
\end{figure}

\section{Methods}


\subsubsection{Time-Frequency Image Transformation.}
\begin{figure}[t]
    \centering
    \includegraphics[width=1\linewidth]{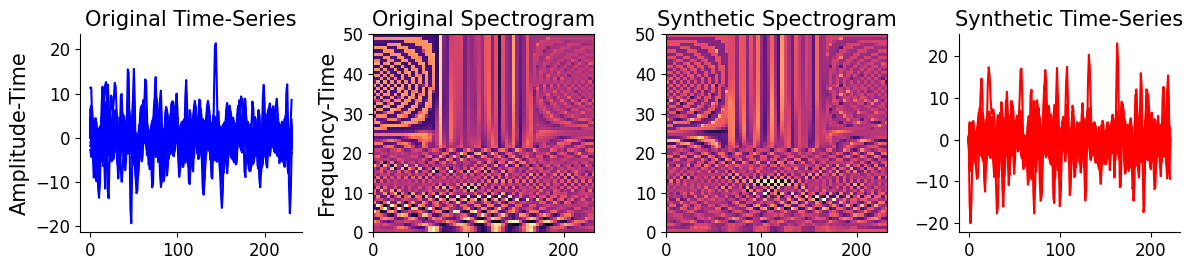}
    \caption{Original vs synthetic BOLD signals and generated normalized spectrograms. Our framework directly generates BOLD signals as opposed to correlation matrices.}
    \label{overview}
\end{figure}

Fig. \ref{fig1} and \ref{overview} provides an overview of our proposed framework. Given high-dimensional fMRI signals from \( S \) subjects, denoted as \( \mathcal{X}_{S} = \{\mathbf{x}_s\}_{s=1}^{S} \), where each subject \( \mathbf{x}_s \in \mathbb{R}^{D \times T} \) consists of \( D \) regions of interest (ROIs) recorded over \( T \) time points, our objective is to learn the underlying real data distribution \( p_{\text{data}}(\mathcal{X}_{S}) \) and generate a synthetic distribution \( p_{\theta}(\mathcal{X}_{S}) \) that is statistically indistinguishable from the real data. 

Unlike conventional time-series generative tasks that operate exclusively in the time domain, our approach transforms fMRI time series into a time-frequency spectrogram image using the windowed Fourier transform (WFT), defined as:
\begin{equation}
\mathbf{X}_s(m,k) = \sum_{n=0}^{N-1} x_s[n + m\, h]\, w[n]\, e^{-j 2\pi \frac{k}{N} n},
\end{equation}
which serves as the input to the generative model. Here, \( n \in \{1, 2, \dots, N\} \) is the local time index within an WFT window of length \( N \); \( m \in \{1, 2, \dots, M\} \) is the window index, where the total number of windows are computed as \( M = \lfloor (T - N)/h \rfloor + 1 \) with signal length \( T \) and hop size \( h \); and \( k \in \{1, 2, \dots, K\} \) is the frequency index, with the number of frequency bins defined as \( K = N/2 + 1 \) since only the first half of the spectrum is retained for real-valued signals. 

Given an univariate input signal \( x_s \in \mathbb{R}^{1 \times T} \), the WFT produces a spectrogram image representation \( \mathbf{X}_s \in \mathbb{R}^{2 \times K \times M} \), where the channels are doubled (2 channels) to store both the real \( \mathbf{X}_s^{\text{Re}}\) and imaginary \( \mathbf{X}_s^{\text{Im}}\) components of the WFT output, and the spatial dimensions (height and width) of the image are determined by the frequency bins \( K \) and time frames \( M \), respectively. We further perform component-wise normalization to accentuate between high and low frequencies. This structured representation allows generative models to process time-series data as images while preserving both spectral and temporal characteristics, enabling robust generative modeling of the fMRI BOLD signals.

To reconstruct the original fMRI time series spectrogram representation, we first unnormalize the real and imaginary components, we merge them to form a complete complex spectrogram \(\hat{\mathbf{X}}_s(m,k) = \mathbf{X}_s^{\text{Re}}(m,k) + j \, \mathbf{X}_s^{\text{Im}}(m,k) \). For each window indexed by $m$ and for local time indices $n$, we compute the inverse fourier transform to reconstruct the fMRI BOLD signals as follows: 
\begin{equation}
x_s[n + m\,h] = \frac{1}{N} \sum_{k=0}^{K-1} \mathbf{X}_s(m,k) \, e^{j\, 2\pi \frac{k}{N} n}.  
\end{equation}

Finally, these segments are recombined using the overlapping method with Hann window to obtain the reconstructed time-domain signal \(\tilde{x_s}\). This process ensures that both the spectral and temporal characteristics of the original fMRI BOLD signals are preserved.

\subsubsection{Classifier-Free Denoising Diffusion.}
Rather than incrementally adding noise through a Markov chain, we employ EDM framework \cite{edm}, which perturb the ground-truth fMRI spectrogram images using a single noise level \( \sigma_{t} \) drawn from a lognormal distribution. At each diffusion step \( t \in \{0, 1, \ldots, T\} \), the images are corrupted according to the perturbation kernel \( p_\theta(x_t \mid x_0) \) which corresponds to \(x_t = x_0 + n_t\),
where the total noise \( n_t = \sigma_t \cdot \epsilon_t \) with 
\( \sigma_t = \exp\left(\eta_t \cdot P_{\text{std}} + P_{\text{mean}}\right) \), \(\eta_t \sim \mathcal{N}(0, 1)\) and \(\epsilon_t \sim \mathcal{N}(0, I)\). Here, \( n_t \) represents a sample from a Gaussian distribution whose variance is modulated by the lognormal factor \(\sigma_t\). This process gradually transforms the fMRI spectrogram images into non-informative Gaussian noise.
The reverse process aims to learn the conditional distribution \(p_\theta(x_{t+1} \mid x_{t})\) to reconstruct the original fMRI spectrogram images. 
In particular, a deep neural network denoiser \(\epsilon_\theta(x_{t}, \sigma_{t})\) is trained to predict the the clean image \(x_0\) from its noisy version \(x_{t}\) at noise level \(\sigma_{t}\) across time steps. Specifically, the noise scheduling is defined as 
\(
\sigma_{t} = \left(\sigma_{\text{max}}^{1/\rho} + \frac{t}{T-1} \Bigl(\sigma_{\text{min}}^{1/\rho} - \sigma_{\text{max}}^{1/\rho}\Bigr)\right)^\rho
\).
Using Heun’s second-order method, the reverse-diffusion update is given by:
\begin{equation}
x_{t+1} = x_{t}' + (\sigma_{t+1} - \sigma_{t}') \cdot \frac{1}{2} \, \left[ \frac{x_{t}' - \epsilon_\theta(x_{t}', \sigma_{t}')}{\sigma_{t}'} + \frac{x_{t+1}' - \epsilon_\theta(x_{t+1}, \sigma_{t+1})}{\sigma_{t+1}} \right],
\end{equation}
where \(\sigma_{t}' = \sigma_{t} + \gamma_{t} \, \sigma_{t}\) and \(x_{t}' = x_{t} + \sqrt{\sigma_{t}'^2 - \sigma_{t}^2} \, \epsilon_{t}' \) is perturbed by noise \(\epsilon_{t} \sim \mathcal{N}(0, S_{\text{noise}}^2 I)\), where both \(\gamma_{t}\) and \(S_{\text{noise}}\) are EDM hyperparameters. By iteratively applying this update from \( t=T \) down to \( t=0 \), a synthetic fMRI spectogram image \( x_0 \sim  p_\theta(x)\) is sampled from the marginal data distribution.



To achieve class-conditioned generation, we apply classifier-free guidance and parameterize the conditional distribution \( p_{\theta}(x_t|c) \) via a conditional noise-prediction model \( \epsilon_\theta(x_t, t, c) \), where $c$ is the subject class. For unconstrained generation, we then input a null token \( \varnothing \) for the unconditional model \( \epsilon_\theta(x_t, t) = \epsilon_\theta(x_t, t, \varnothing) \).
Concurrently, these noise-prediction models are trained to minimize the discrepancy between the true noise \( \epsilon \) and the predicted noise \( \epsilon_\theta \) across timesteps, via the noise-matching objective
\(\mathcal{L}(\theta) = \mathbb{E}_{x_t, \epsilon, t, c} \left[ \left\| \epsilon - \epsilon_\theta \right\|^2 \right]\). 
The unconditional and conditional models are jointly trained by randomly setting \( c \) to the unconditional class identifier \( \varnothing \) with some discrete probability \( p_{\varnothing}\). 

\section{Experiments and Results}
\textbf{Data Acquisition and Pre-processing.}
We preprocessed the dataset from REST-meta-MDD Consortium database \cite{restmdddb} using Data Processing Assistant for Resting-State fMRI (DPARSF) \cite{dprasf}. This resting-state fMRI dataset consist of 250 healthy controls (HC) and 227 Major Depressive Disorder (MDD) patients. The data were acquired using a Siemens (Tim Trio 3T) scanner (TR/TE = 2000/30 ms, 3mm slice thickness). The brain was parcellated into 116 region-of-interests (ROIs), including cortical and subcortical areas, and the mean time series of 232 time points for each ROI was extracted using the Automated Anatomical Labeling (AAL) atlas. 

\subsubsection{State-of-the-art Baselines.}
i) Quality Evaluation: Our proposed T2I-Diff model is compared against six time-series generative model baselines. These baselines include generative adversarial networks (GANs) and diffusion models such as CoT-GAN \cite{cotgan}, DiffTime \cite{difftime}, DiffWave \cite{diffwave}, TimeVAE \cite{timevae} and Diffusion-TS \cite{diffusionts}.
ii) Classification Score: We further evaluate the performance of our proposed model with a special designed classifier for brain connectvity \cite{brainnetcnn}. The baselines include 1D-Deep Convolutional GAN (1D-DCGAN) \cite{dcgan} and Wasserstein GAN Gradient Policy(WGAN-GP) \cite{wgangp}. 

\subsubsection{Implementation Details.}
i) Connectivity Network Construction:
The functional connectivity of brain networks is constructed using Ledoit-Wolf (LDW) regularized shrinkage covariance estimator to keep the strongest $\tau = 40\%$ connections and set other connections to zeros. This results in FCs of size 116x116 for each subject. 
ii) T2I-Diff Training: The proposed T2I-Diff framework generates the fMRI signals corresponding to the subjects' condition (HC and MDD). The BrainNetCNN classifier then discriminates between the HC and MDD subjects. We train the T2I-Diff via an Adam optimizer using a learning rate of $3e^{-4}$ for 1000 epochs. For all experiments, we use 18 diffusion sampling steps and experiment in image sizes of 8×8, 16×16, 32×32, and 64×64. 
iii) Data Augmentation and Classifier Training: The trained T2I-Diff is used to augment fMRI signals of 1×, 2× and 3× on top of the real fMRI signals. For our classifier, the L2 regularization weight decay from $10^{-8}$ to  $10^{-2}$ , scheduler learning rate reduce factor from 0.1 to 0.9, batch size from 5 to 16 same as in \cite{GR-SPD-GAN}. These model hyperparameters are selected based on a 5-fold stratified cross-validation.

\begin{table}[t]
    \centering
    
    \caption{Comparison of fMRI signal generation quality with SOTA time-series generation and our proposed T2I-Diff.}
    \label{fMRI_generation}
    
    \resizebox{\textwidth}{!}{%
    \begin{tabular}{l|c|c|c|c|c|c|c}
    \toprule
     & CoT-GAN & DiffTime & DiffWave & TimeVAE & TimeGAN & Diffusion-TS & \textbf{T2I-Diff} \\
    \midrule
    Context-FID 
    & 7.813$\pm$.550 
    & 0.340$\pm$.015 
    & 0.244$\pm$.018 
    & 14.449$\pm$.969 
    & 0.126$\pm$.002 
    & 0.105$\pm$.006 
    & 1.384$\pm$.107 \\
    \midrule
    Correlational 
    & 26.824$\pm$.449 
    & 1.501$\pm$.048 
    & 3.927$\pm$.049 
    & 17.296$\pm$.526 
    & 23.502$\pm$.039 
    & 1.411$\pm$.042 
    & 4.121$\pm$.094 \\
    \midrule
    Discriminative 
    & 0.492$\pm$.018 
    & 0.245$\pm$.051 
    & 0.402$\pm$.029 
    & 0.476$\pm$.044 
    & 0.484$\pm$.042 
    & 0.167$\pm$.023 
    & 0.400$\pm$.059 \\
    \midrule
    Predictive 
    & 0.185$\pm$.003 
    & 0.100$\pm$.000 
    & 0.101$\pm$.000 
    & 0.113$\pm$.003 
    & 0.126$\pm$.002 
    & 0.099$\pm$.000 
    & 0.102$\pm$.001 \\
    \bottomrule
    \end{tabular}
    }
\end{table}

\begin{table}[t]
    \centering
    
    \caption{Ablation study of short-term and long-term fMRI signal generations across sequence lengths 24, 64, 128 and 256.}
    \label{ablation}
    
    \resizebox{\textwidth}{!}{%
    \begin{tabular}{c|c|c|c||c|c|c|c}
    \toprule
     \textbf{24} 
     & \textbf{64} & \textbf{128} & \textbf{256} 
     & \textbf{24} & \textbf{64} & \textbf{128} & \textbf{256} \\
    \midrule
     \multicolumn{4}{c||}{\textbf{Context-FID}} & \multicolumn{4}{c}{\textbf{Correlation}} \\
    \midrule
    1.384$\pm$0.107
    & 7.141$\pm$0.789 
    & 10.655$\pm$1.001 
    & 1.661$\pm$0.146 
    & 4.121$\pm$0.094 
    & 4.737$\pm$0.136 
    & 3.525$\pm$0.101 
    & 4.124$\pm$0.243 
    \\
    \midrule
    \multicolumn{4}{c||}{\textbf{Discriminative}} & \multicolumn{4}{c}{\textbf{Predictive}} \\
        \midrule
    0.400$\pm$0.059
    & 0.280$\pm$0.268 
    & 0.176$\pm$0.090 
    & 0.360$\pm$0.164 
    & 0.102$\pm$0.001 
    & 0.100$\pm$0.001 
    & 0.098$\pm$0.001 
    & 0.095$\pm$0.002 \\
    \bottomrule
    \end{tabular}
    }
\end{table}

\subsection{Overall Performance}


      





We first trained our diffusion models unconstrained to produce similar outputs. Then, we followed the standard setting for the evaluation of time series generation from Diffusion-TS \cite{diffusionts}.
The purpose of our experiments is to demonstrate the effectiveness of the proposed time-frequency representation and classifier-free diffusion models in achieving a trade-off between FID and predictive scores, comparable to other time-series generative models. Rather than solely aiming for state-of-the-art sample quality metrics on these benchmarks, our primary goal is to validate the ability of our approach to model complex spatiotemporal patterns and excel in capturing conditional distribution over time.

\subsubsection{fMRI Signal Generation Quality.}
Table \ref{fMRI_generation} compares our T2I-Diff framework with state-of-the-art time-series generative models across four evaluation metrics for a time-series generation length of 24. T2I-Diff demonstrates competitive overall performance, surpassing strong time-series generation models such as CoT-GAN, DiffWave, TimeGAN, and TimeVAE. Notably, the low discriminative and predictive scores suggest that T2I-Diff effectively generates synthetic fMRI BOLD signals with in-distribution temporal patterns and spectral features that may not be fully captured in the time domain. 
By leveraging the WFT image transform, the diffusion model not only preserves global contextual information but also models the complex spatiotemporal dynamics inherent in fMRI signals. 
While T2I-Diff does not achieve the highest overall scores, our results demonstrate that the time-series-to-image generative framework is well-suited for modeling high-dimensional, complex fMRI time series.

\subsubsection{Ablation Study.}
Table \ref{ablation} assess the impact of sequence length on model performance, we conducted ablation studies on time-series generation lengths of 64, 128, and 256 to contrast the sequence length of 24.
Shorter sequences exhibit a low context-fid score and high predictive score, but their lower predictive score suggests a limited ability to capture temporal dynamics due to insufficient contextual information. Mid-length sequences (64 and 128) show improved context-fid scores but struggle with increased complexity. The longest sequence (256) achieves the best context-fid and predictive scores, effectively capturing both global and fine-grained features. 

\subsubsection{Classification Score.}

\begin{table}[t]
    \centering
    \caption{Classification performance of different classifiers trained on the ground-truth data and increasing amount of augmented time series data using our proposed model.}
    \label{fMRI_classification}
    \resizebox{\textwidth}{!}{%
    \begin{tabular}{l|l|c|c|c|c|c}
    \toprule
    Method
    &Train Set
    &Accuracy 
    &Recall
    &Precision 
    &F1-Score
    &ROC  \\
    \midrule
    W/O Augmentation
        & Real & 58.90 ± 2.98 & 58.90 ± 2.98 & 59.56 ± 2.74 & 58.39 ± 3.09 &  59.00 ± 2.56 \\
    \midrule
    \multirow{3}{*}{1D-DCGAN}
      & Real + Synth 1× & 62.94 ± 2.01 & 62.94 ± 2.01 & 63.43 ± 2.20 & 62.23 ± 2.68 & 62.71 ± 2.26 \\
      & Real + Synth 2× & 65.04 ± 2.02 & 65.04 ± 2.02 & 66.35 ± 2.13 & 64.12 ± 2.10 & 64.74 ± 2.08 \\
      & Real + Synth 3× & 58.21 ± 2.98 & 58.21 ± 2.98 & 55.70 ± 6.58 & 52.86 ± 4.14 & 57.38 ± 3.11 \\
    \midrule
    \multirow{3}{*}{WGAN-GP}
      & Real + Synth 1× & 66.02 ± 4.25 & 66.02 ± 4.25 & 66.22 ± 4.24 & 65.93 ± 4.20 & 65.95 ± 4.13 \\
      & Real + Synth 2× & 64.76 ± 4.25 & 64.76 ± 4.25 & 65.67 ± 4.08 & 64.23 ± 4.52 & 64.73 ± 4.14 \\
      & Real + Synth 3× & 64.56 ± 3.18 & 64.56 ± 3.18 & 64.78 ± 3.17 & 64.38 ± 3.15 & 64.41 ± 3.08 \\
    \midrule
    \multirow{3}{*}{\textbf{T2I-Diff (Ours)}}
      & Real + Synth 1× & 66.87 ± 3.22 & 66.87 ± 3.22 & 67.06 ± 3.34 & 66.83 ± 3.21 & 67.26 ± 6.00 \\
      & Real + Synth 2× & 65.41 ± 2.37 & 65.41 ± 2.37 & 66.30 ± 1.67 & 64.73 ± 2.80 & 65.75 ± 3.22 \\
      & Real + Synth 3× & 66.03 ± 1.75 & 66.03 ± 1.75 & 66.50 ± 1.32 & 65.85 ± 1.82 & 66.58 ± 5.33 \\
    \bottomrule
    \end{tabular}
    }
\end{table}

To validate the fidelity of the generated samples, we evaluate the classification performance of BrainNetCNN and compare it against ID-DCGAN and WGAN-GP on our proposed fMRI dataset. Here, we maintain the original sequence length rather than truncating, as supported by the performance of long sequence generation in Table \ref{fMRI_generation}. Table \ref{fMRI_classification} presents the classification results on the 5-fold cross-validation test set. Notably, T2I-Diff achieves the highest accuracy in the 1× data augmentation setting. Furthermore, the results indicate that as the augmentation range increases, our model exhibits lower variance, suggesting that the synthetic data generalizes well across varying augmentation levels while maintaining consistent classification performance. 
The results highlight that T2I-Diff not only enhances the diversity of synthetic fMRI samples but also effectively captures and preserves critical structural and functional patterns for downstream tasks such as brain disorder classification.

\begin{figure}[t]
    \centering
    \includegraphics[width=1\linewidth]{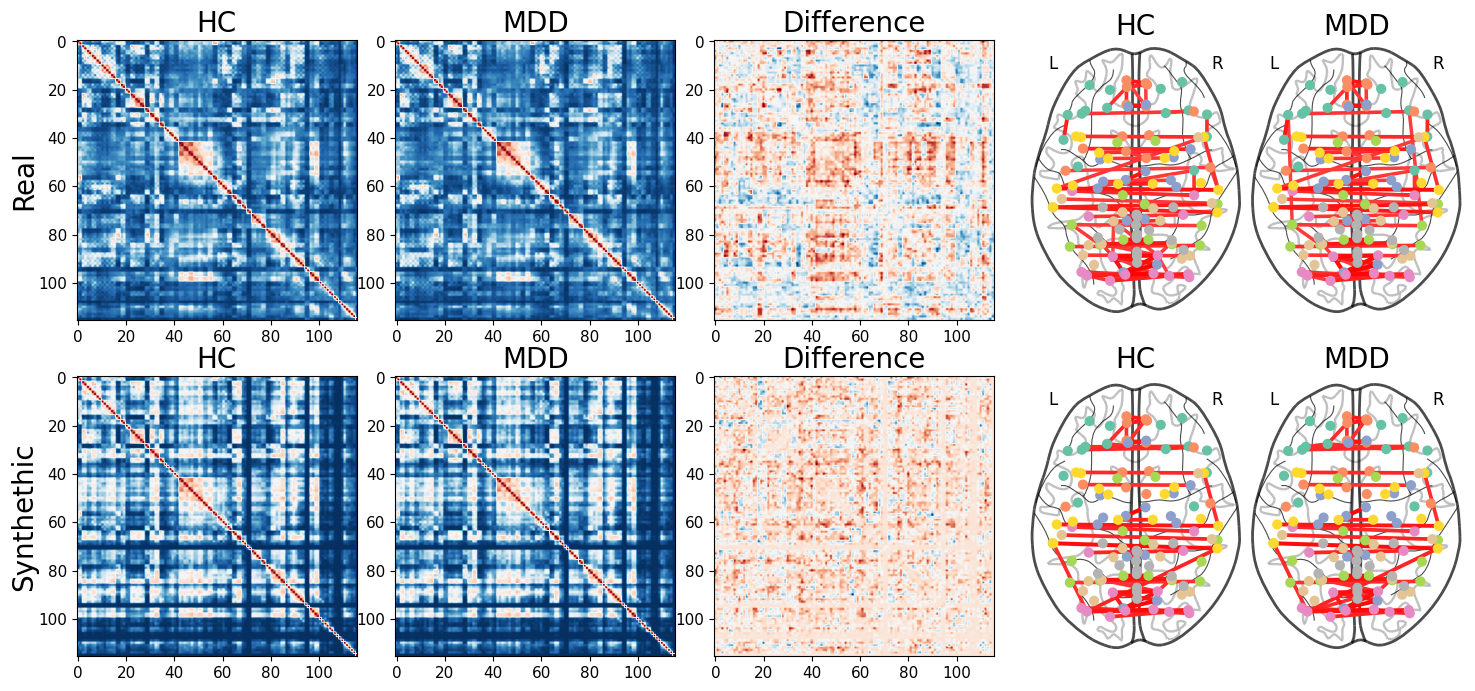}
    \caption{Real and synthethic (left) connectivity pattern and (right) brain networks.}
    \label{visualization}
\end{figure}
\subsection{Functional Connectivity (FC) Visualization}
To further evaluate the quality of generated data, we derived functional connectivity (FC) measures from the synthetic fMRI BOLD signals to assess differences between healthy controls (HC) and major depressive disorder (MDD) patients. We computed the average connectivity using a threshold of 0.6 to highlight significant connections. Our analysis reveals that the synthetic FC closely aligns with the functional changes observed in the real FC distribution. Furthermore, connectogram comparisons between HC and MDD in both real and synthetic FC data indicate a reduction in connectivity within the left superior frontal gyrus (FrontalSupL) and decreased connectivity between the left middle frontal gyrus (FrontalMidL) and the anterior cingulate cortex (CingulumAntL). The results suggest impaired cognitive functions, such as difficulties in decision-making and emotion regulation, reinforcing the biological plausibility of the generated data.

\section{Conclusions and Future Work}
In this paper, we propose T2I-Diff, which effectively captures both temporal dynamics and spectral evolution underlying the ground-truth data distribution for accurate brain signal generation. For future work, we aim to further validate MDD classification using graph-based deep learning models \cite{tew2024kans,tew2025st}. Furthermore, we plan to incorporate energy-based models to identify out-of-distribution (OOD) patterns in brain spectogram associated with neurological disorders \cite{loo2025learning}.


\begin{credits}
\subsubsection{\ackname} This work was supported by the Advanced Computing Platform (ACP), Monash University Malaysia. The work of Junn Yong Loo is supported by Monash University under the SIT Collaborative Research Seed Grants 2024 I-M010-SED-000242.

\subsubsection{\discintname}
The authors have no competing interests to declare that are relevant to the content of this article.
\end{credits}

%
%
%
\bibliographystyle{splncs04}
\bibliography{Paper-3042.bib}
%




\end{document}